\documentclass[conference]{IEEEtran}
\usepackage[normalem]{ulem}
\usepackage{mathptmx} 

\usepackage{fancyhdr}
\usepackage[normalem]{ulem}
\usepackage[hyphens]{url}
\usepackage[sort,nocompress]{cite}
\usepackage[final]{microtype}
\usepackage[keeplastbox]{flushend}
\usepackage[bookmarks=true,breaklinks=true,letterpaper=true,colorlinks,linkcolor=black,citecolor=blue,urlcolor=black]{hyperref}
\usepackage{amsmath}
\usepackage{amssymb}
\usepackage{tabularx}

\usepackage{graphicx}
\usepackage{listings}
\usepackage{setspace}
\usepackage{subfigure}
\usepackage[most]{tcolorbox}
\newcommand{\insertFigure}[2]{
    \begin{figure}[t]
\setlength{\abovecaptionskip}{0pt}
\setlength{\belowcaptionskip}{0pt}
        \centering
        \includegraphics[width=\linewidth]{figs/#1.pdf}
        \caption{\small #2}
	\vspace{2mm}
        \label{fig:#1}
    \end{figure}
}

\newcommand{\insertWideFigure}[2]{

    \begin{figure*}[h]
\setlength{\abovecaptionskip}{0pt}
\setlength{\belowcaptionskip}{0pt}
        \centering
        \includegraphics[width=\textwidth]{figs/#1.pdf}
	\vspace{-4mm}
        \caption{\small #2}
	\vspace{-2mm}
        \label{fig:#1}
    \end{figure*}

}

\newcommand{\squishlist}{
 \begin{list}{$\bullet$}
  { \setlength{\itemsep}{0pt}
     \setlength{\parsep}{3pt}
     \setlength{\topsep}{3pt}
     \setlength{\partopsep}{0pt}
     \setlength{\leftmargin}{1.5em}
     \setlength{\labelwidth}{1em}
     \setlength{\labelsep}{0.5em} } }

\newcommand{\squishlisttwo}{
 \begin{list}{$\bullet$}
  { \setlength{\itemsep}{0pt}
     \setlength{\parsep}{0pt}
    \setlength{\topsep}{0pt}
    \setlength{\partopsep}{0pt}
    \setlength{\leftmargin}{2em}
    \setlength{\labelwidth}{1.5em}
    \setlength{\labelsep}{0.5em} } }

\newcommand{\squishend}{
  \end{list}  }

\newcommand{\recnet}{\textsc{AIrchitect}}

\title{
    \recnet{}: Learning Custom Architecture Design and Mapping Space
}

\begin{document}
\author{
  Ananda Samajdar\\
  \textit{Georgia Tech}\\
  \textit{Atlanta, GA}\\
  \texttt{anandsamajdar@gatech.edu}
  \and
  Jan Moritz Joseph\\
  \textit{RWTH Aachen University}\\
  \textit{Aachen, Germany}\\
  \and
  Matthew Denton\\
  \textit{Georgia Tech}\\
  \textit{Atlanta, GA}\\
  \and
  Tushar Krishna\\
  \textit{Georgia Tech}\\
  \textit{Atlanta, GA}\\
  
}
\maketitle
\pagestyle{plain}

\author{
  Ananda Samajdar\\
  \textit{Georgia Tech}\\
  \textit{Atlanta, GA}\\
  \texttt{anandsamajdar@gatech.edu}
  \and
  Jan Moritz Joseph\\
  \textit{RWTH Aachen University}\\
  \textit{Aachen, Germany}\\
  \and
  Matthew Denton\\
  \textit{Georgia Tech}\\
  \textit{Atlanta, GA}\\
  \and
  Tushar Krishna\\
  \textit{Georgia Tech}\\
  \textit{Atlanta, GA}\\
  
}
\begin{abstract}

Design space exploration is an important but costly step involved in the design/deployment of custom architectures to squeeze out maximum possible performance and energy efficiency.
Conventionally, optimizations require iterative sampling of the design space using simulation or heuristic tools. 
In this paper we investigate the possibility of learning the optimization task using machine learning (ML) and hence using the learnt model to predict optimal parameters for the design and mapping space of custom architectures, bypassing any exploration step.
We use three case studies involving the optimal array design, SRAM buffer sizing, mapping, and schedule determination for systolic-array-based custom architecture design and mapping space.
%
We perform systematic design-aware and statistical analysis of the optimization space for our case studies and highlight the patterns in the design space.
We formulate the architecture design and mapping as a ML problem that allows us to leverage existing ML models for training and inference.
We design and train a custom network architecture called \recnet{}, which is capable of learning the architecture design space with as high as 94.3\% test accuracy and predicting optimal configurations which achieve on average (GeoMean) of 99.9\% the best possible performance on a test dataset with $10^5$ GEMM workloads. 
\end{abstract}
\section{Introduction}
\label{sec:intro}


Custom architectures, in the form of accelerators, are arguably the de facto solution to create high performance hardware in the face of diminishing benefits from semiconductor device scaling. 
Among the many areas where custom architectures helped to meet the ever-increasing demand for compute performance, ML acceleration is probably the most noticeable example.
As ML models start to become ubiquitous and more complex across various applications,
their sheer demand for computing power and their inherent parallelism renders them too expensive and inefficient for general-purpose architectures; simultaneously creating a perfect case to reap the benefits of custom-designed accelerators \cite{eyeriss}. 
Unsurprisingly ML accelerators are now used across all form factors from data centers to mobile phones, providing record-breaking performance and energy efficiency \cite{tpu, tpu-edge}. 

Custom architecture design is a data-driven process. 
In a typical design cycle, the application characteristics are carefully studied for understanding possible optimizations, software bottlenecks, and computation/memory access patterns.
Then multiple iterations of design space exploration (DSE) are performed for evaluating the costs of possible architecture choices via simulations or other cost models \cite{DSE-theory, scalesim-ispass}. 
Finally, based on the data obtained from DSE and applicable constraints, an implementation choice is made to get the best performance and efficiency.

In this paper, we demonstrate that the same data-driven decision-making can be learned by a ML model for the architecture design task. 
To get insights into the patterns, dimensionality, feasibility of learning the architecture design and mapping space we undertake three case studies.
The first case study concerns the problem of finding the optimal shape and mapping strategy of a monolithic systolic array when presented with a general matrix multiply (GEMM) workload and design constraints.
The second deals with sizing of SRAM buffer dimensions in a monolithic systolic array, when queried with the information about the buffer size limits, interface bandwidth, structure and mapping of the compute array, and the workload dimensions.
The third case study explores the scheduling space of optimally running GEMM workloads on systolic arrays with different configurations.

\insertFigure{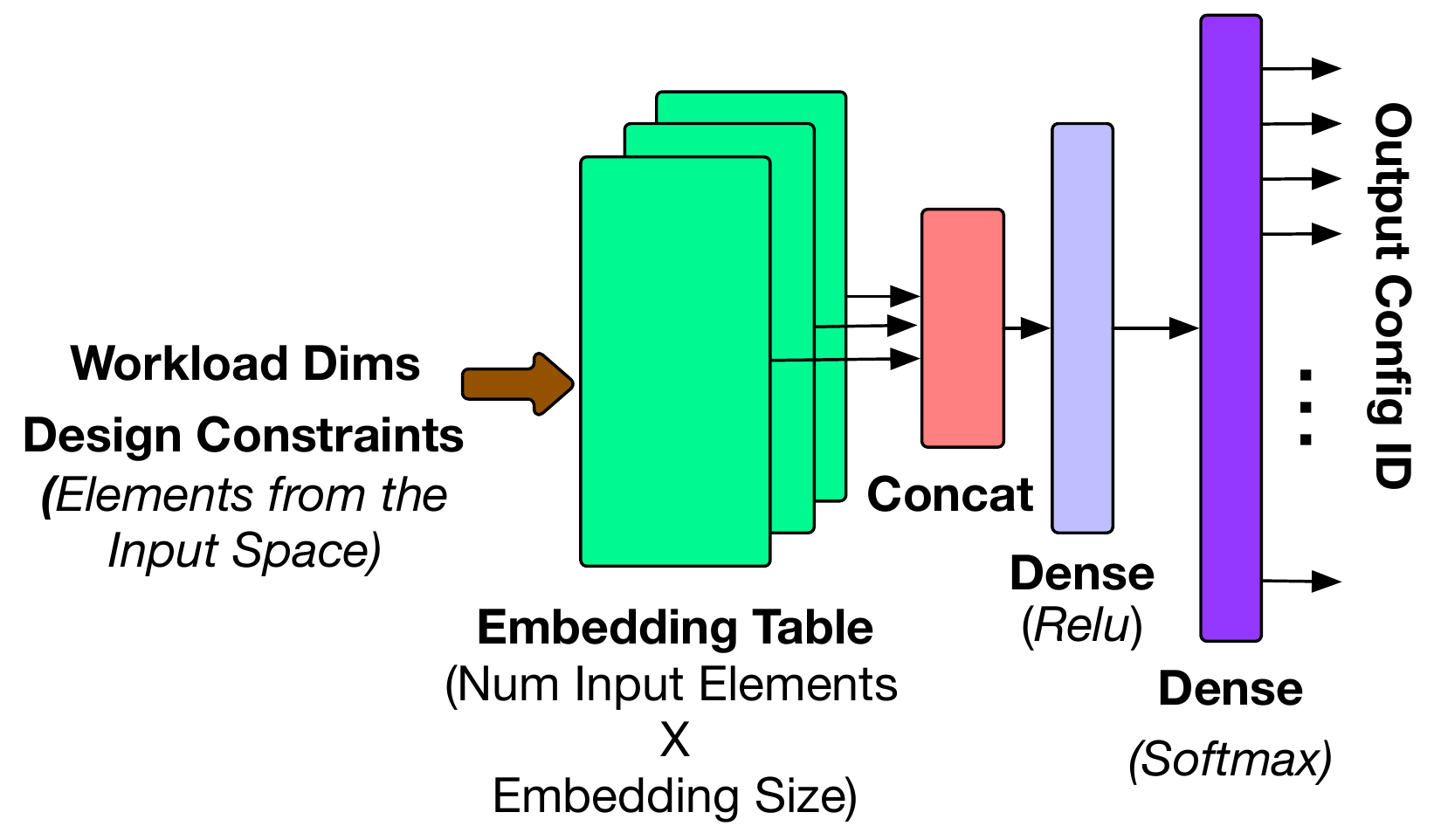}{General structure of the proposed class of recommendation neural networks (\recnet).}

With the help of our case studies, we systematically examine the 
optimization space of design and mapping to understand patterns and relationships between input parameters and optimal points, which can potentially facilitate learning of the space.
Our analysis demonstrates that the architecture design and mapping space is learnable, at least within the limits of our use cases.
We demonstrate that by formulating the optimization prediction as a ML classification, we achieve high degrees of accuracy in determining the optimal design parameters using a learned model. 
We design and evaluate a custom class of neural network architectures called \recnet{} that is capable of learning the design space to get a prediction accuracy as high as 94.3\% on a test dataset of $10^5$ datapoints.

\textit{To the best of our knowledge, this work is one of the first that learns the design space to \textbf{generalize} it}. 
Most prior works leveraging ML for accelerator DSE~\cite{autotvm, reagen2017case,ahn2019reinforcement,confuciux,gamma} focus on performing the search faster.
Learning the design space enables constant time prediction of the optima. Such a feature has the potential to significantly speed up the conventional architecture design cycles.
For instance, custom architecture design, by its very definition is tightly coupled with an application or an algorithm, which enables extracting orders of magnitude performance and energy efficiency as compared to general-purpose hardware.
However, this tight dependence means that architecture design needs to be refreshed in tandem with changes to the application. 
%
For deep learning accelerator architectures, over the years, optimization objectives shifted from supporting GEMMs, to native convolutions~\cite{eyeriss,maeri}, and more recently to attention layers \cite{kung-systolic,Centaur}.
For each such design cycle, an expensive DSE is needed to be performed repeatedly to find the architecture parameters corresponding to the new optima.
A ML model similar to our proposal will significantly reduce the cost of such design iteration by condensing down the search for optimal configurations to constant time.

In summary, the following are our contributions:

\squishlist
    \item{
        \textit{We present systematic, design-aware (\autoref{sec:design-aware-analysis}), and statistical studies(\autoref{subsec:statistical-analysis}) on the design and mapping space of custom architectures via three case studies, demonstrating the feasibility of learning these spaces.}
    }
    \item{
        \textit{
            We formulate traditional DSE tasks as ML classification/recommendation tasks by "quantizing" the optimization space and structuring the input and the output spaces. (\autoref{sec:dataset-generation-analysis})
        This abstraction enables different optimization tasks to be converted into learning problems, solvable by the use of existing ML classification models.}
    }
    \item{
        \textit{
            We proposes and evaluate \recnet{} (\autoref{fig:recnet-generic}), a custom-designed neural recommendation network capable of learning the DSE space to a high degree of prediction accuracy (\autoref{sec:airchitect}). The predicted optimal configurations generated by \recnet{} achieve 99.99\% of the best achievable performance on a test dataset of $10^5$ workloads when used for our case studies.
        }
    }
\squishend


The paper is organized as follows.
\autoref{sec:casestudy} defines each of our case studies and illustrate the complexity of the problem for finding the optima for a single query.
In \autoref{sec:design-aware-analysis}, we generate multiple datapoints for each of the case studies and perform a design-aware analysis to identify underlying patterns across the datapoints which help with learning the space of optimum configurations.
In \autoref{sec:dataset-generation-analysis}, we 
cast the design-optimization as a learning problem, generate a training dataset, and perform statistical analysis.
Finally we use both off-the-shelf and our custom designed ML model \recnet{} (\autoref{sec:airchitect}) to learn the design space and perform predictions.
In \autoref{sec:related-works} we examine the existing literature, before concluding in \autoref{sec:conclusion}.

\section{Case Studies}
\label{sec:casestudy}


\insertFigure{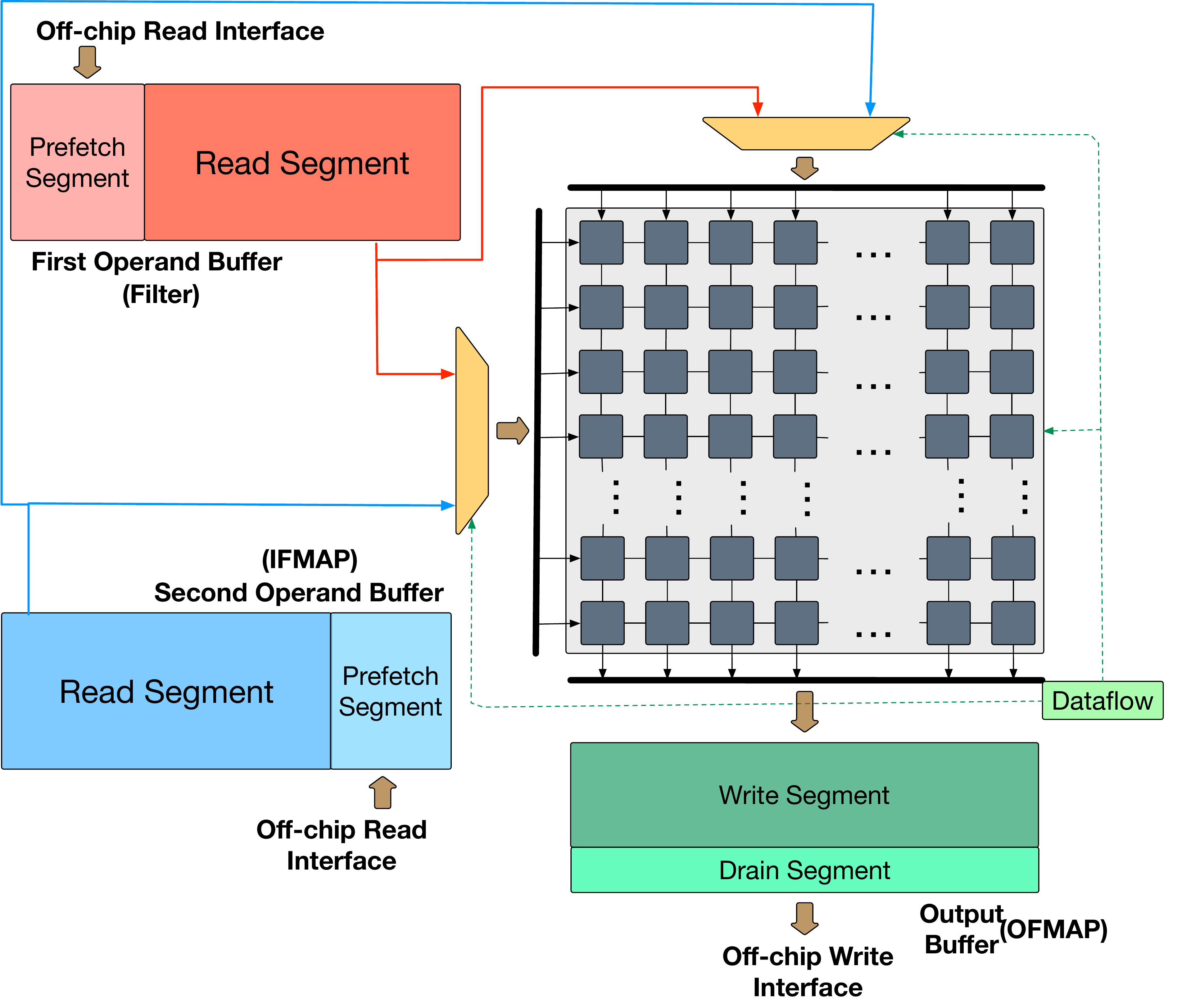}{
    Monolithic systolic array template with configurable array dimensions, choice of dataflow, and configurable buffer sizes
}

In this work, we have contrasting requirements of choosing the design goal, which is complex enough to be challenging for naive design approaches but is feasible for human experts to reason about and validate the experimental results.
We choose the problem of efficient systolic arrays-based GEMM accelerator design and mapping as our design-target for a ML model to learn.
The simple construction and the easy scalability of systolic arrays creates a large design-space controlled via a few input parameters. Furthermore, a large abundance of literature for these structures fosters understanding the design space and examining the outcomes of the learning algorithms.

\subsection{Background}

\textbf{Systolic Array.} Systolic arrays are highly efficient structures for dense matrix multiplication and naturally have been adapted in many commercial and academic proposals for DNN acceleration~\cite{tpu, xdnn, nvidia-tensor-core}.
\autoref{fig:SystolicHighLevel} depicts the schematic template for the target of our design task.

\textbf{Dataflow.} 
There are three distinct strategies of mapping
compute, aka dataflows, onto the systolic array named Output
Stationary (OS), Weight Stationary (WS), and Input Stationary (IS)~\cite{eyeriss}. The “stationarity” of a
given dataflow is determined by the tensor whose element
is not moved (i.e. stationary) for the maximum duration of
time throughout the computation. Although many different
dataflows exist for spatial arrays~\cite{maestro}, we only consider true systolic dataflows that only use local (neighbor-to-neighbor) communication in this work.
af
\subsection{Case Study Description}
In this paper, we target three design-spaces / case studies:

\textbf{ -- Case Study 1: Array and Dataflow Prediction.} This case study focuses on the compute array in \autoref{fig:SystolicHighLevel}. The array has two sets of design parameters we wish to learn and recommend for a given workload (i.e., DNN layer): (i) shape of the array, determined by choosing the rows and columns, and (ii) dataflow.

\textbf{ -- Case Study 2: Buffer Size Prediction.} Given a compute array of a fixed size and shape, this case study 
focuses on sizing the SRAM buffers around it. As shown in \autoref{fig:SystolicHighLevel}, the array is supported by three buffers. Two of the buffers prefetch and store the input operands (Input Feature map (IFMAP) and Filter in Convolution neural network terminology\cite{eyeriss}), which are fed into the array from the top or left edges as determined by the choice of dataflow. 
The third buffer acts as temporary storage for the generated output elements (Output feature map (OFMAP)) or partial sums until they are drained. 

\textbf{ -- Case Study 3: Multi-array Scheduling.} Given several heterogeneous arrays (with their own fixed sizes and memories), shown later in \autoref{fig:SchedulerHighLevel}, this case study focuses on scheduling multiple independent layers across them.


We motivate each case study in more detail next. 
We use SCALE-Sim\cite{scalesim-ispass} and some in house simulation tools to generate the data points (see \autoref{sec:dataset-generation-analysis}).


%

\subsection{Compute Design Space}
\label{subsec:casestudy-arr-df}

For our systolic-array-based accelerator, the parameters in the compute design space are the shape of the array, and the mapping strategy or dataflow. 
In GEMM workloads, owing to the large parallelism and ease of scaling, the more computation elements (multiply-and-accumulate or MAC units) that are allocated to a problem the better. 
However, just providing the extra compute resources is not enough. The mapping and the shape of the systolic array determine the fraction of these compute resources that are utilized for work. 
Determining the best array shape and mapping has been widely covered in the existing literature. 
However, for the sake of our discussion,  we perform some simple experiments to depict the complexity of the problem.

First, we depict the effect of the shape of the array when running the first layer of ResNet-18\cite{resnet} in \autoref{fig:arr_df_variation_resnet18}.
\autoref{fig:arr_df_variation_resnet18}(a) shows three plots, each for the dataflow Output Stationary (OS), Weight Stationary (WS), and Input Stationary (IS), respectively.
In each of these charts, we plot the runtime obtained for running the given layer for different array shapes possible when using $2^9$ MAC units, with each array having at least 4 rows or columns. 
In \autoref{fig:arr_df_variation_resnet18}(b), we depict the result for a similar analysis on this layer (first layer in ResNet-18) for the various array shapes when $2^{15}$ MAC units are used.
First, we observe from this chart is that we can have orders of magnitude in difference in the runtime between a poor configuration and the optimal one.
For example, for $2^{15}$ MACs using OS dataflow, using a shape $4\times8192$ takes $100\times$ more cycle than when using the optimal configuration of $512\times64$.
Second, we see that for a given shape of the array, the array utilization correlates with better runtimes. This is logical as, the higher the utilization is, the more parallelism can be exploited resulting in higher performance. 
However, utilization is not a sufficient condition for determining optimality. 
For example, when using OS dataflow with $2^9$ MAC units, all array shapes but 1 has full utilization ($8\times64$ to $128\times4$); however the runtimes among the configurations vary with the optimal configuration ($16\times32$) taking almost half the time than $128\times4$.
When considering the case of $2^{15}$ units and IS dataflow, the optimal configuration ($64\times512$) has a lower utilization than 100\% obtained by $4\times8192$ and still achieves about a tenth in runtime.

\insertFigure{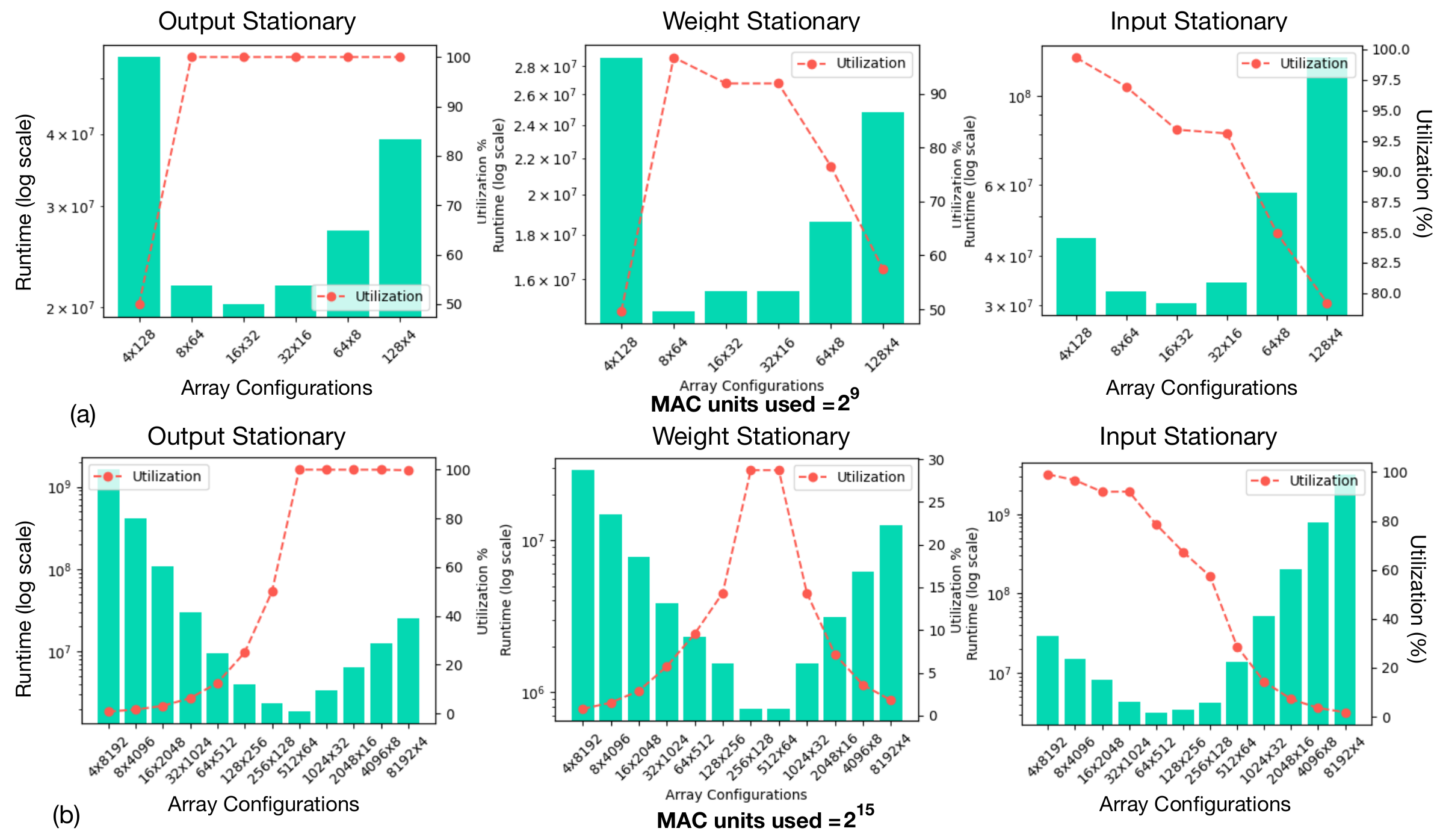}{
    Variation of runtime and utilization (red dotted line) when using different array shapes and dataflows for (a) $2^9$ MAC units, and (b) $2^{15}$ MAC units for first layer in ResNet18
}

\insertFigure{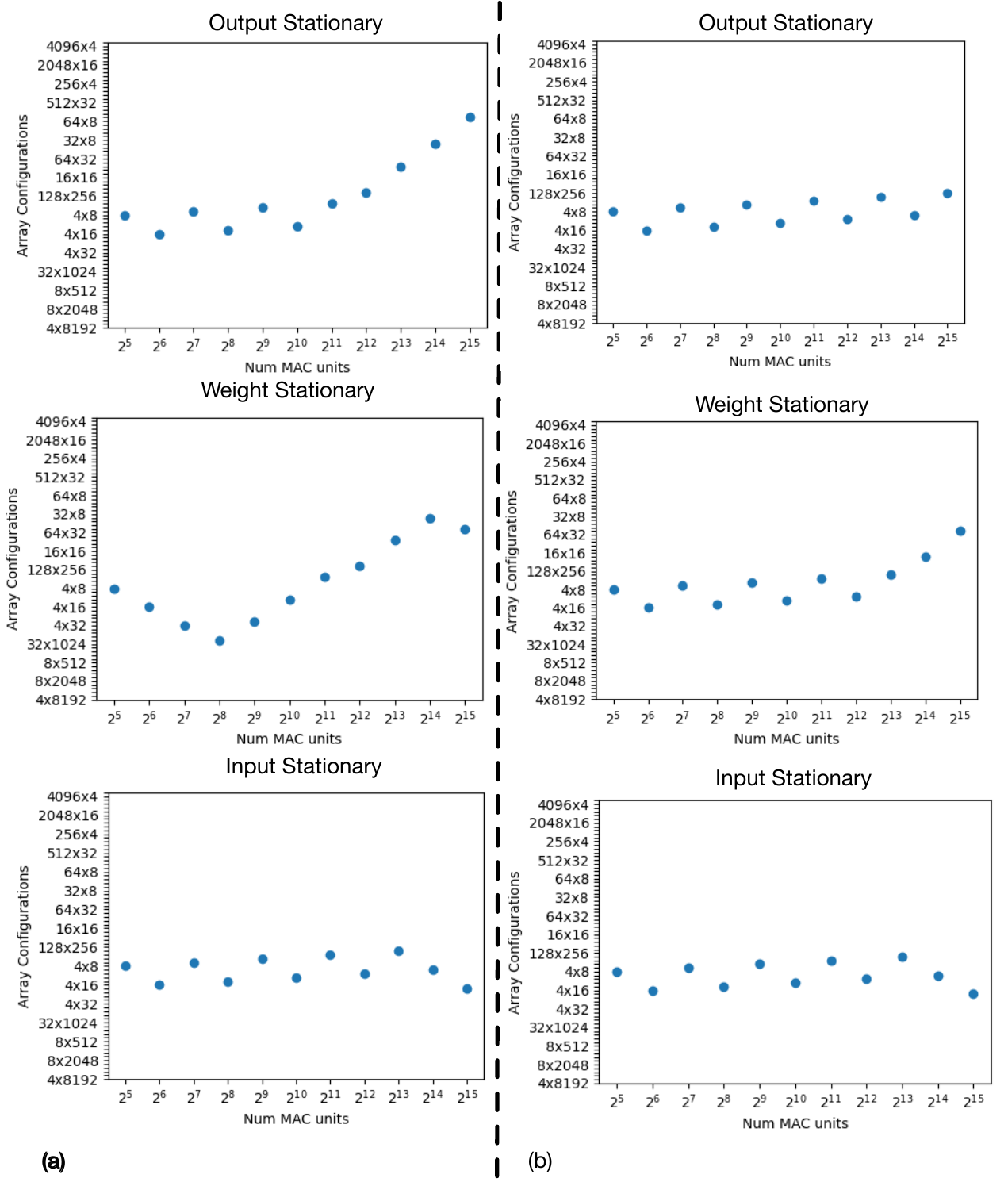}{
    The variation of optimal array dimensions as the number of compute units vary from $2^{5}$ to $2^{15}$ for (a) the first layer and (b) the eighth layer in Resnet18 network for different dataflows.
}

We note that even with the same workload and dataflow, the optimal shape of the array changes with the design constraint itself.
In \autoref{fig:aspect_ratio_variation_resnet18}, we plot the array shape of the optimal point when the number of MAC units used changes from $2^5$ to $2^{15}$ for the first layer and eighth layer of ResNet18\cite{resnet}.
The y-axis of the plot depicts the various configurations sorted by aspect ratios (the ratio of rows to columns), thus the axis with fewer rows and more columns (short-fat) are near the x-axis, the ones with comparable rows and cols (balanced) are in the center, while the ones with more rows and fewer columns (tall-skinny) are at the top. 
We observe that for a small number of MAC units, the balanced array configurations are preferred for both the layers and dataflows. However, as the number of MAC units is changed. the OS mappings prefer a "tall-skinny" configuration for the first layer. 
The WS mappings initially tend towards a "short-fat" configuration but then lean towards a "tall-skinny" configuration, whereas the IS mapping does no show any such trend in both layers. 

These studies show that, as the workloads and design constraints change, the optimal architecture and mapping configurations vary. 
However, the optimal architecture parameters do not follow any immediate pattern that can be approximated and used to make fast predictions using simple modeling methods.

\vspace{-1mm}
\subsection{SRAM Buffer Design Space}
\label{subsec:casestudy-buf-design}

 \vspace{-1mm}

\insertFigure{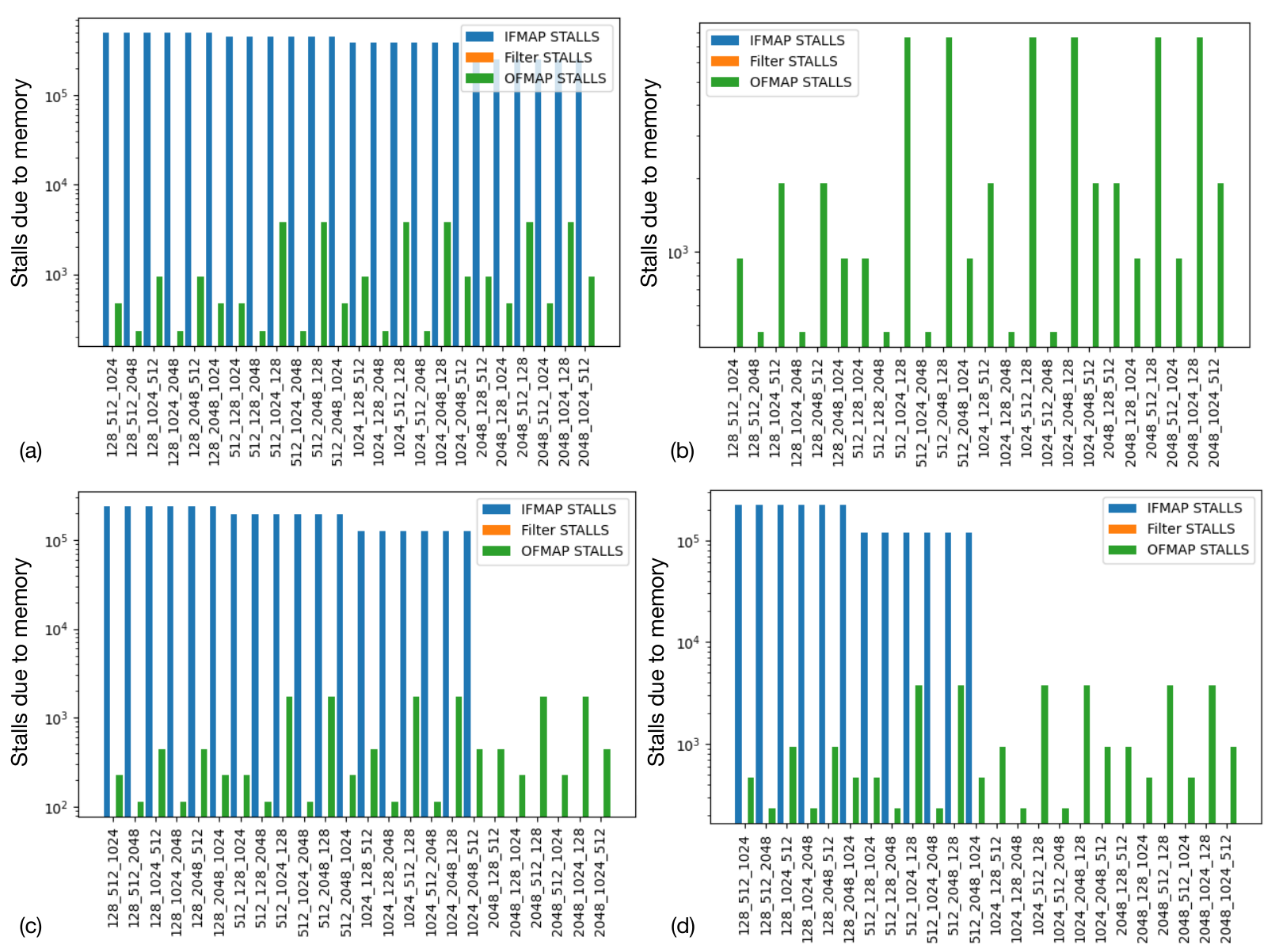}{
Variation of stalls encountered due to memory limitation vs memory size categorized per buffer type for  GoogleNet's\cite{googlenet} second layer when using (a) OS dataflow, 32x32 array, and interface BW of 50 bytes per cycle; (b) IS dataflow, 32x32 array, 50 bytes/cycle BW (c) OS dataflow, 32x32 array, 100 bytes/cycle BW (d) OS dataflow, 8x128 array, 50 bytes/cycle BW}


Sizing the memories properly is yet another important problem that architects need to address when building an efficient accelerator. 
When designing an accelerator, there is typically an upper bound placed on the real estate available for compute, buffering, other components in the chip. 
For the systolic-array-based accelerator depicted in \autoref{fig:SystolicHighLevel}, the capacities for buffers holding the three operands need to be allocated out of this budget. 
It is well known that the relative sizes of the buffers depend on the dimensions of the operands themselves.
This task becomes challenging and non-intuitive given the fact that several extraneous factors like mapping/dataflow, other architecture parameters like the array dimensions, external bandwidth, etc also determine the relative sizes. 

We illustrate this by plotting the cost of improper allocation in \autoref{fig:mem_stall_variation_googlenet}.
In this figure, we show the stall obtained due to improper memory allocation on the second layer of GoogleNet, when running on arrays with $2^{10}$ MAC units with various dataflows and external bandwidth. 
For this experiment, we only allow the buffer sizes among 128 KB, 512 KB, 1024 KB, and 2048 KB and estimate the stalls encountered.
In \autoref{fig:mem_stall_variation_googlenet}(a) we depict the stalls encountered when the layer is run on a $32\times 32$ array with OS dataflow, with each of the buffers backed by a 50 words/cycle interface bandwidth link.
For this case, we observe that allocating the maximum amount of capacity for the buffer containing Input Feature Map (IFMAP) buffer leads to minimum stalls. The buffer containing Filter operands can be set to the lowest size as all sizes lead to zero stalls, while the rest of the capacity can be allocated for storing the outputs.
In \autoref{fig:mem_stall_variation_googlenet}(b), we just change the dataflow from OS to IS and notice a significant change in the stall behavior.
In this case, the only buffers that account for non-zero stalls are the ones responsible for holding the output buffer elements (ie the Output feature map or OFMAP). 
Consequently, the optimal sizing allocates minimum buffer sizes for the input matrices and the maximum to the buffer holding OFMAP elements. 

Varying other architecture parameters like the interface bandwidth and the dimensions of the array also change the optima.
When the interface bandwidth is changed to 100 words/cycle while keeping the same dataflow (OS) and the array dimensions ($32\times 32$), we observe that allocating 2048 KB to IFMAP buffer leads to zero stalls from IFMAP transfers, as shown in \autoref{fig:mem_stall_variation_googlenet}(c). 
In this case, given the stalls are then only encountered from the OFMAP buffer, the configuration with 1048 KB allocated to this buffer turns out to be optimum. 
When using array dimensions of $8\times 128$, the optimal is obtained by allocating 1024 KB for the IFMAP and 2048 KB for the OFMAP lead to the best configuration (see fig \autoref{fig:mem_stall_variation_googlenet}(d)).
The lack of any apparent pattern and the dependence of the optimal buffer sizing on the workload dimension and architecture parameters limits the alternatives of optimization to iterative search-based methods.

\vspace{-2mm}
\subsection{Multi Array Scheduling Space}
\label{subsec:casestudy-sched-space}

 \vspace{-1mm}
 

\insertFigure{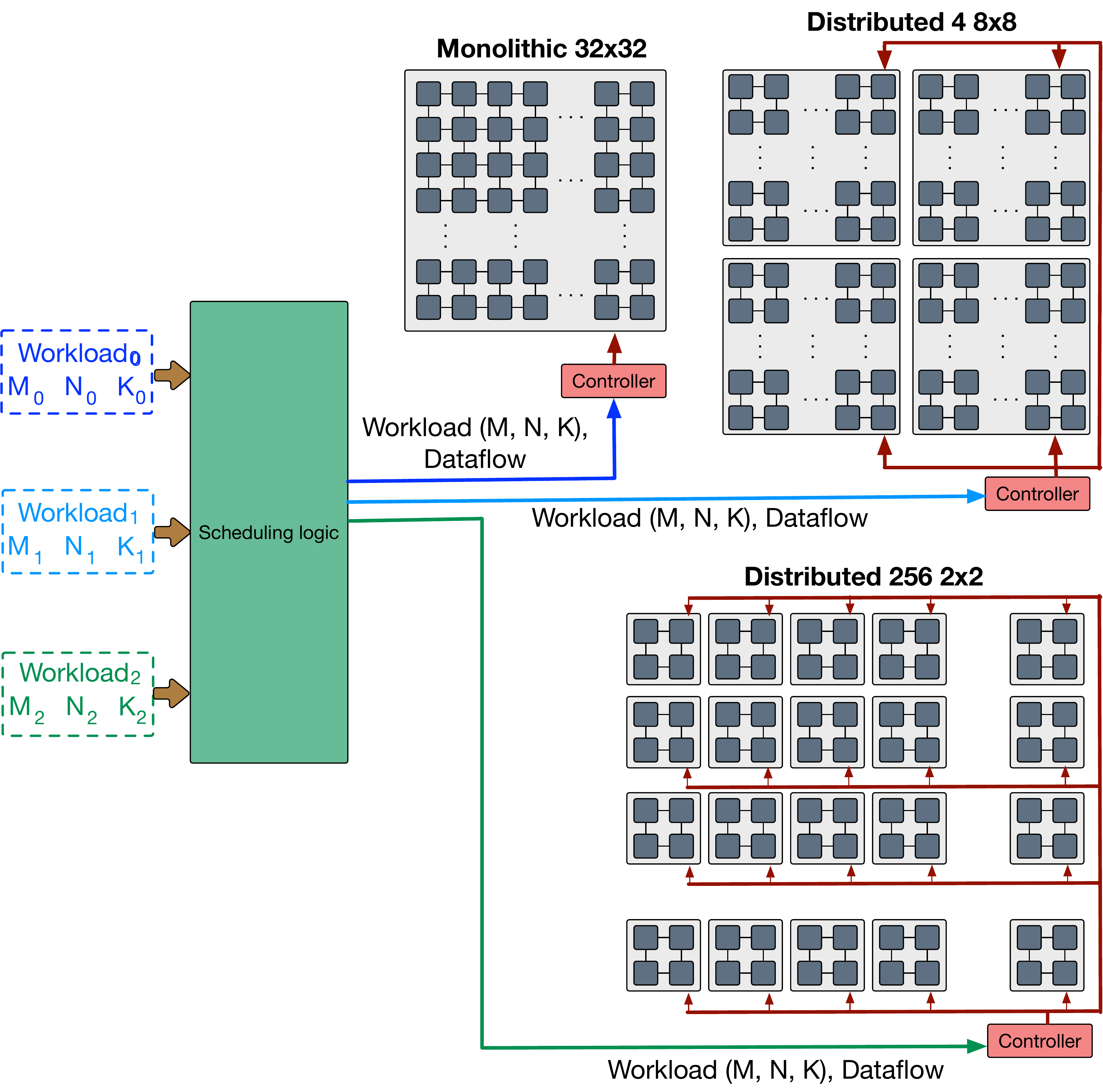}{Schematic description of the multi-array scheduling case study}

\insertFigure{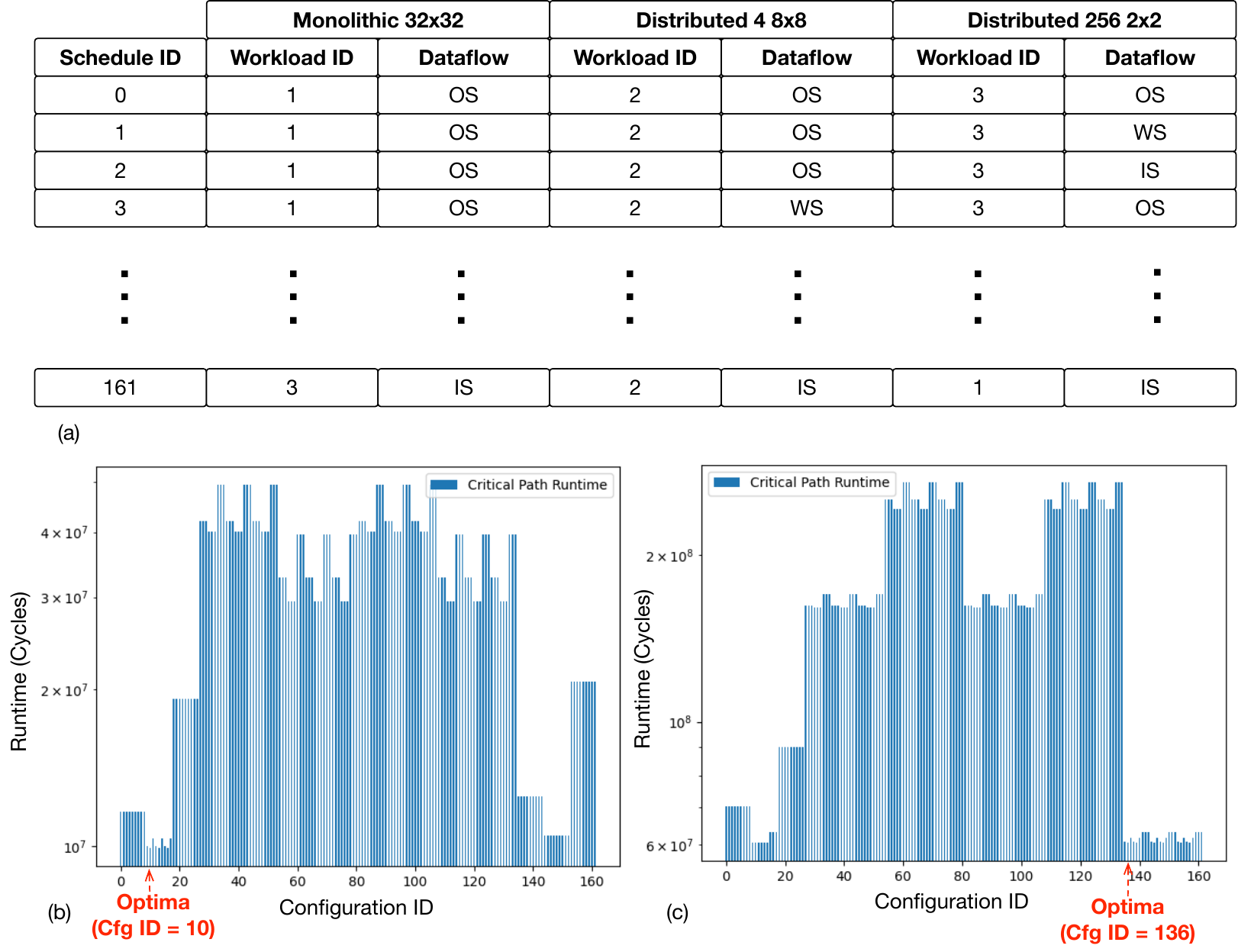}{
    (a) Table of scheduler IDs and corresponding configuration
    (b) Runtimes of the critical path for the first layer in GoogleNet, YoloTiny, and Alexnet
    (c) Runtimes of the critical path for the third layer in GoogleNet, YoloTiny, and Alexnet
}

In our third case study, we examine if the scheduling space of a heterogeneous accelerator can be learned.
Although the scheduling problem can come in many forms, for our experiments the specific problem we address of matching a fixed number of GEMM workloads to an equal number of systolic array-based distinct compute units, such that the overall runtime is minimized. 
\autoref{fig:SchedulerHighLevel} helps describe the problem qualitatively.
In this particular example, we have three compute units, one a monolithic $32\times32$ systolic array, a distributed collection of 4 $8\times8$ systolic arrays, and a fine-grained distributed accelerator with 256 $2\times2$ systolic arrays. 
As evident by the example, the distinct compute units can be a single or a distributed collection of arrays, working on one GEMM workload at a time, 
For each such unit, the dataflow is tunable at runtime.
Hence, the task of the scheduler is to find the match among each workload to the corresponding array, when queried with three workload dimensions, and to chose the dataflows such that the overall runtime is the lowest.

This task is equivalent to the collection of several tasks like (i) identifying the workload on the critical path, (ii) finding the best compute unit and corresponding dataflow to run the critical path workload the fastest, (iii) performing the same analysis for the second, third, etc slowest network in case of a tie among various possible scheduling combinations such that sum of time taken on all the compute arrays can be minimized.

The table in \autoref{fig:sched_critical_path_runtime}(a) depicts the number of schedules possible for this simple example. 
\autoref{fig:sched_critical_path_runtime}(b) depicts the runtime of the network on the critical path for all the possible schedules for the first layers in GoogleNet, YoloTiny, and Alexnet. 
The abrupt variation of runtimes for the neighboring schedules suggests that search is the practical way of determining to find the optima in this search space. 
Also, among the various workloads, the optimal schedule varies widely. 
For example, in \autoref{fig:sched_critical_path_runtime}(c) we show a similar chart for the critical path runtime of third layers in the previous networks. 
The optimal schedule, in this case, is the one corresponding to entry 136 of the table in \autoref{fig:sched_critical_path_runtime}(a), as compared to entry 10 in the previous figure.
As with the previous case studies, predicting an optimal schedule is also challenging due to the lack of any apparent pattern, making it an interesting candidate for learning.

\vspace{-2mm}

\section{Design aware analysis}
\label{sec:design-aware-analysis}

\vspace{-2mm}


In this section, we analyze the datasets for our three design targets from an architect's perspective to understand the design space and to highlight the trends and generalizations in the design space.
We call this design aware analysis since we use the knowledge of the design space and the input architecture and workload parameter to understand the observed patterns.

\subsection{Array shape and dataflow}
\label{subsec:design-aware-analysis-arr-df}
 
 \vspace{-1mm}

\insertFigure{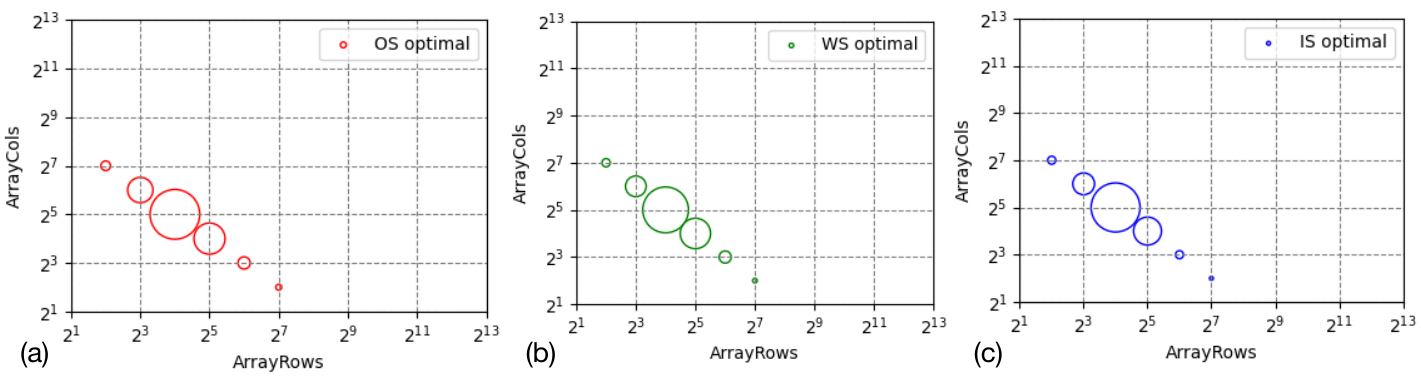}{Relative frequency of optimal array dimensions obtained for GEMM workloads using (a) Output Stationary (b) Weight Stationary, and (c) Input Stationary dataflows}

%

\insertFigure{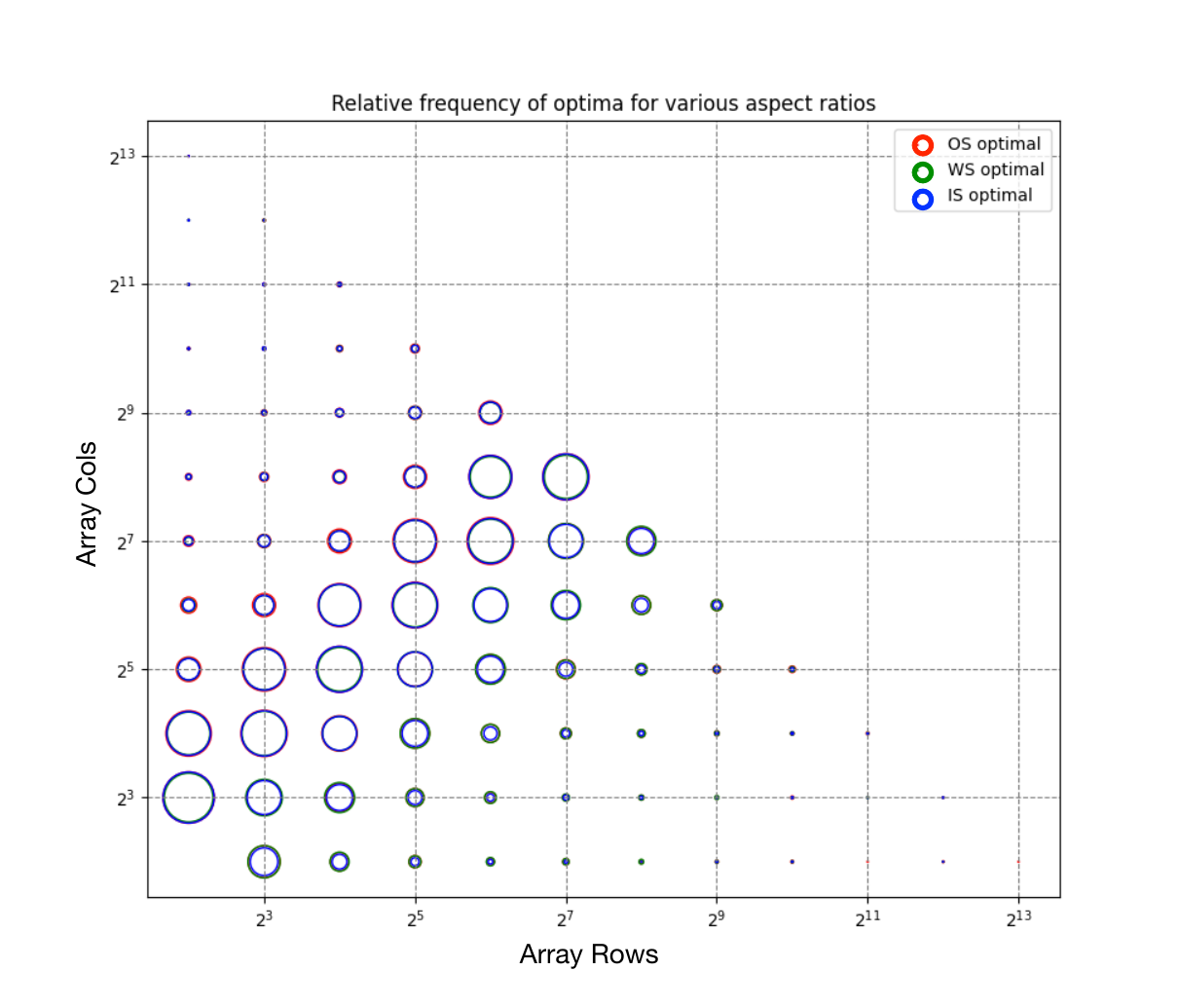}{
The pattern of optimal aspect ratio and optimal dataflow obtained for GEMM workloads on systolic arrays with $2^8$ to $2^{15}$ MACs. 
}

In \autoref{subsec:casestudy-arr-df}, we discussed the complexity of determining the optimal array shape and dataflow for a given GEMM workload and resource constraints. 
However, we know that human experts can generalize to trim the search space or to find heuristics for making searches quicker with the knowledge acquired from experience.
To further understand the design space and identify patterns for the optimum architecture and mapping configurations, we study the optimal design points obtained and we look into the solutions obtained by a search for a handful of workloads.

First, we examine the space of optimal array dimensions (height and width)
obtained in \autoref{fig:ArrDf-AspectRatio-Scatter-2e9} when the design constraint (i.e., the maximum number of MAC units) is fixed to $2^9$.
This chart is generated by searching the optimal for $10^4$ GEMM workloads, obtained by the sampling the operand dimensions (M, N, and K) from the distribution depicted in \autoref{fig:mnk_distribution_sched_space_growth_comb}(a) (see \autoref{sec:dataset-generation-analysis} for details on dataset generation). 
In \autoref{fig:ArrDf-AspectRatio-Scatter-2e9}(a) we show the optimal array dimensions for workloads that obtain the best runtime when using OS dataflow. 
The radius of the marker at each array dimension (row/col) captures the relative frequency at which the said dimension is deemed optimal.
\autoref{fig:ArrDf-AspectRatio-Scatter-2e9}(b,c) captures similar information for the workloads for which the optimal dataflow is WS and IS dataflow, respectively.

In \autoref{fig:ArrDf-AspectRatio-Scatter-sweep_comb} we capture the relative frequency of optimal array dimensions for different values of maximum MAC units varied from $2^{5}$ to $2^{15}$. 
In each 'circle' of the figure, the optimal configurations for each dataflow (similar to the plot in \autoref{fig:ArrDf-AspectRatio-Scatter-2e9}) is combined and presented using a different color. 
We observe that first the optimal array configurations always use the maximum number of MAC units, which is expected. 
Another observation, however, is that the optimal array configuration does not show any dependence on the optimal dataflow. 

\autoref{fig:ArrDf-AspectRatio-Scatter-sweep_comb} also helps in identifying the underlying pattern by combining the plots. 
The following observations can be made from this figure. 
\textit{First}, we see a clear pattern that, the most frequent array configurations are the ones with are near square shape but have twice as many columns as rows, unlike the conventional approach of choosing square configurations \cite{tpu, xdnn}.
The square configuration of the array although optimal for a reasonable amount of workloads, does not work for most of them.
\textit{Second}, all of the array configurations are found to be optimal for at least one workload. Thus, none of the possible array dimensions can be summarily rejected from the search space to find the global optimum. 
\textit{Third}, there is no apparent pattern that can be used for determining the optimal dataflow given the information about the optimal array configurations.

\insertWideFigure{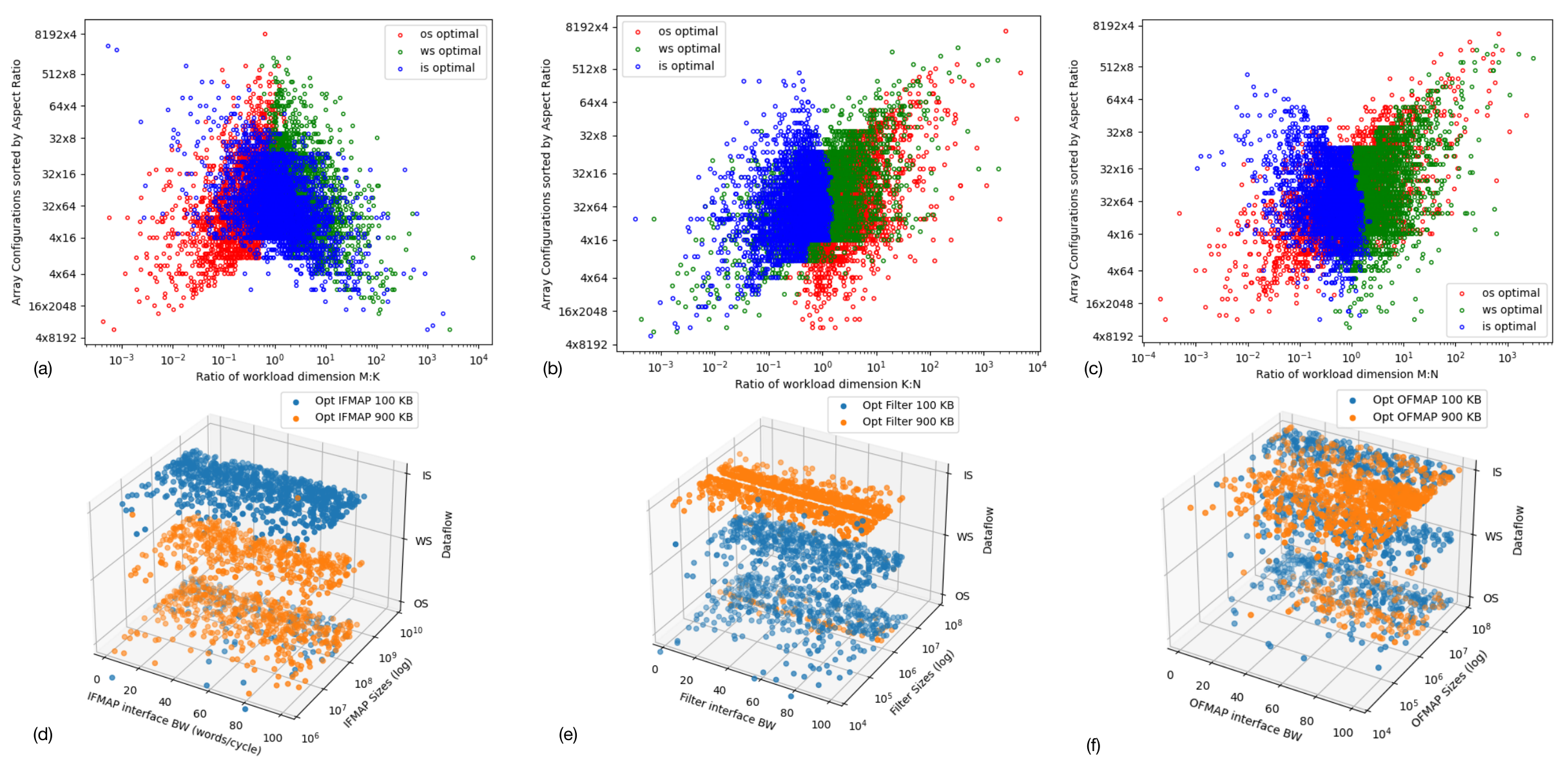}{
The correlation of optimal array dimension obtained (y-axis) and the matrix shape in terms of aspect ratio (x-axis) of (a) Input operand matrix (M$\times$K) (IFMAP), (b) input operand matrix (K$\times$N) (Filter), and (c) Output matrix (M$\times$N) (OFMAP). The colors of the markers indicate the optimal dataflow obtained, highlighting the pattern in the design space1.
The relation of optimal buffer sizes of (d) IFMAP operand buffer, (e) Filter operand buffer, and (f) OFMAP buffer, with operand sizes, interface bandwidth, and dataflow.
}

To understand the variation of optimal dataflow predicted, we study the design space for the shape of the operand matrices as depicted in \autoref{fig:design_aware_comb}(a-c).
For example, in \autoref{fig:design_aware_comb}(a) the x-axis captures the ratio of height and weight of the first operand (M:K) capturing the shape of this matrix. 
The y-axis captures all the possible dimensions of the arrays using MAC units that range from $2^5$ to $2^{15}$ in powers of 2.
Each data point represents the optimal array configuration and dataflow obtained by a GEMM workload, where the dataflow is represented by colors.
The plots in \autoref{fig:design_aware_comb}(b, c) depict similar distributions wrt the shape of the second operand matrix ($K\times N$) and the output matrix ($M\times N$).
One can observe from \autoref{fig:design_aware_comb}(a) that distinguishing between OS and WS is possible just from the shape of the first operand matrix. However, determining either over IS dataflow is difficult as there exists a significant overlap. %
Similarly, \autoref{fig:design_aware_comb}(c) depicts that the dimensions of the output operand matrix can help chose between WS and IS, while \autoref{fig:design_aware_comb}(b) shows the same for IS over OS depending on the shape of the second operand. 

\subsection{SRAM Buffer Sizing}
\label{subsec:design-aware-analysis-mem}

 \vspace{-1mm}


Similar to the task of determining optimal array shape and dataflow for a given GEMM workload, given a constrained real estate the task of determining the optimal sizes of on-chip SRAM buffers is also challenging. 
In \autoref{subsec:casestudy-buf-design} we discussed the challenges involved in this task for a couple of example use cases. 
In a similar vein to our analysis in \autoref{subsec:design-aware-analysis-arr-df}, we try to understand if any generalizations or patterns exist in the design space of the on-chip memory buffer. 
For this study, we consider the optimal buffer dimensions in a systolic array, obtained by searching for $10^4$ distinct GEMM workloads running on randomly chosen array dimensions between $2^4$ to $2^{18}$, interface bandwidth between 1 to 100, and dataflow among OS, WS, and IS. 
For the sake of simplicity, we limit the search space to buffer sizes from 100KB to 900KB, changed in increments of 100KB.
The objective of the search algorithm is to find the sizes of the three buffers simultaneously such that the overall stall cycles are minimized, and in case of multiple configurations with the same cost, the one with cumulatively minimum size.

\autoref{fig:design_aware_comb}(d-f) depicts the correlation of the optimal buffer sizes, with the input parameters, interface bandwidth, the size of the workload matrices, and the dataflow.
For the sake of clarity in all the figures, we plot the points corresponding to the minimum and maximum obtainable sizes.
In \autoref{fig:design_aware_comb}(d) we observe that, when using IS dataflow, small IFMAP buffers are the optimum. 
For WS dataflow, allocating the maximum buffer size appears to be compulsory. 
This observation makes sense when we consider the reuse patterns. 
In IS dataflow, the IFMAP matrix has the maximum reuse, whereas the elements of the filter operand matrix are streamed into the array, thus allowing a small buffer for the input to be sufficient. 
On the other extreme the WS dataflow requires the inputs to be streamed into the array, which requires a larger buffer size to lower the interface bandwidth pressure.
Similar observation can be made in \autoref{fig:design_aware_comb}(e) where the optimal sizes for the filter operand buffers are depicted. 
For this buffer, the smaller buffer dimensions are predominant when WS is used since the filter matrix has the maximum reuse. 
Similarly, larger buffer dimensions are obtained when IS is used. 
However, for the OS dataflow, where both IFMAP and Filter operands are streamed, larger buffers are favored by the IFMAP operand, while small buffers are common for the Filter operand. 
This trend can be explained by comparing the distribution of operand sizes of the matrices. 
The IFMAP matrix sizes vary from $10^8$ to $10^9$ bytes while the Filter operands vary from $10^5$ to $10^7$ bytes making the input matrices about two orders of magnitude larger. 
When allocating from a fixed budget, it is perhaps optimal to allocate maximum memory to the larger matrix.

\autoref{fig:design_aware_comb}(c) depicts the similar plot for the output buffer. 
In this chart, we observe that across the dataflow, small buffer sizes are optimum in general. This makes sense by considering the fact the OFMAP buffer does not contribute to extracting operand reuse and instead acts as temporary storage to hide the slow external interface. 
However, we notice that for large interface bandwidths and relatively small OFMAP sizes, larger OFMAP buffers are preferred. 
This observation can be explained by the fact that the optimal sizes are determined for all three buffers together. 
Furthermore, in this dataset we use the same interface bandwidth for all the buffers.
When the bandwidths of the input interfaces are comparatively high, while the operand sizes are small (M and N) unless the K dimension is extraordinarily large, small input buffers are sufficient to obtain stall-free or low stall operation. 
Thus the stalls encountered from data draining contribute to a performance loss.
At the same time, given there is a 'surplus' of allowable buffer, the size of the output can be increase to reduce the overall stalls encountered.

\vspace{-2mm}
\subsection{Multi-array Scheduling}
\label{subsec:design-aware-analysis-sched}

This case study is a mapping problem similar to the optimal dataflow selection shown in \autoref{fig:design_aware_comb}(a-c). 
The patterns observed in \autoref{fig:design_aware_comb}(a-c) are almost similar to this case. 
For the sake of brevity, we directly show the statistical analysis for this case study in \autoref{fig:dataset_distr_pca_comb}(e)

\vspace{-2mm}
\section{Learning Architecture and Mapping Space}
\label{sec:dataset-generation-analysis}

\vspace{-2mm}

Our analysis in the previous section shows that the systolic-array-based accelerator design and mapping space possesses high-level patterns that indicate that it is perhaps possible to predict design parameters given that the distribution of the data is internalized. %
In this section, we discuss systematically structuring the learning problem, dataset generation, and briefly perform statistical analysis on the generated datasets.

\subsection{Design Optimization as Learning Problem}
\label{subsec:ml-problem=formulation}
\vspace{-1mm}
The first step towards learning the observed patterns is to formulate the search-based optimization problems involved in our case study to machine learning-based regression or classification problem setting.
Empirically we found classification tends to be a better fit for learning architecture and mapping optimization. 
A naive approach for employing classification for predicting architecture parameters is to independently predict each design parameter. 
However, there are a couple of problems with this approach. 
First, a separate model needs to be trained for each design parameter, which can get easily get out of hand when dealing with a system with significant complexity. 
Second, the parameters are often inter-dependent, for example, the sizes of memory buffers in the systolic array, and different models might fail to capture the interdependence if trained separately.
Motivated by these factors, we formulate the problem into a recommendation setting, where a bunch of parameters is clubbed into bins, and the trained model is expected to recommend the bin corresponding to the optimal parameters. 
This formulation helps us leverage the existing classifier models developed by the machine learning community and also helps us reason about the problem systematically.

\subsection{Dataset Generation}
\label{subsec:dataset-generation}

\vspace{-1mm}


\insertFigure{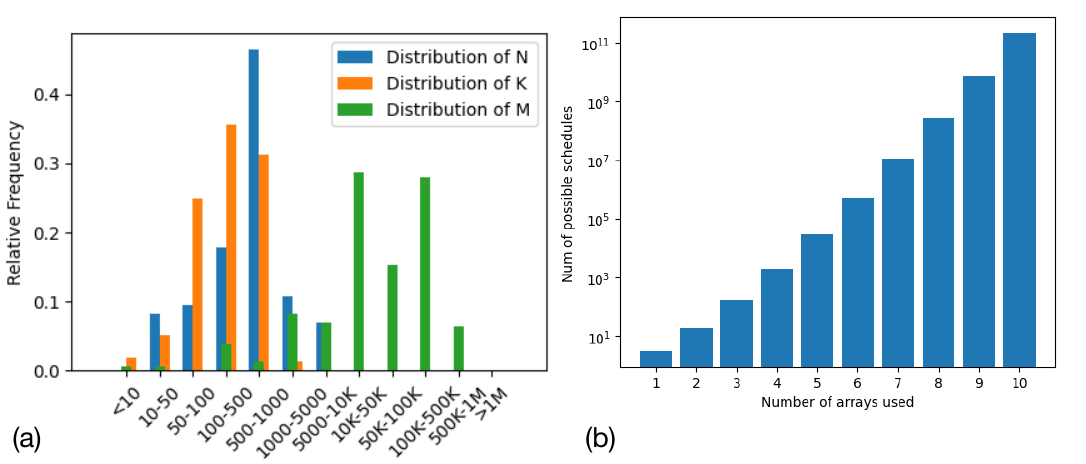}{
    (a) Chart showing the distribution of operand matrix dimensions for GEMM operations involved in layers of popular neural networks
    (b) Growth of the scheduling space
}

\insertFigure{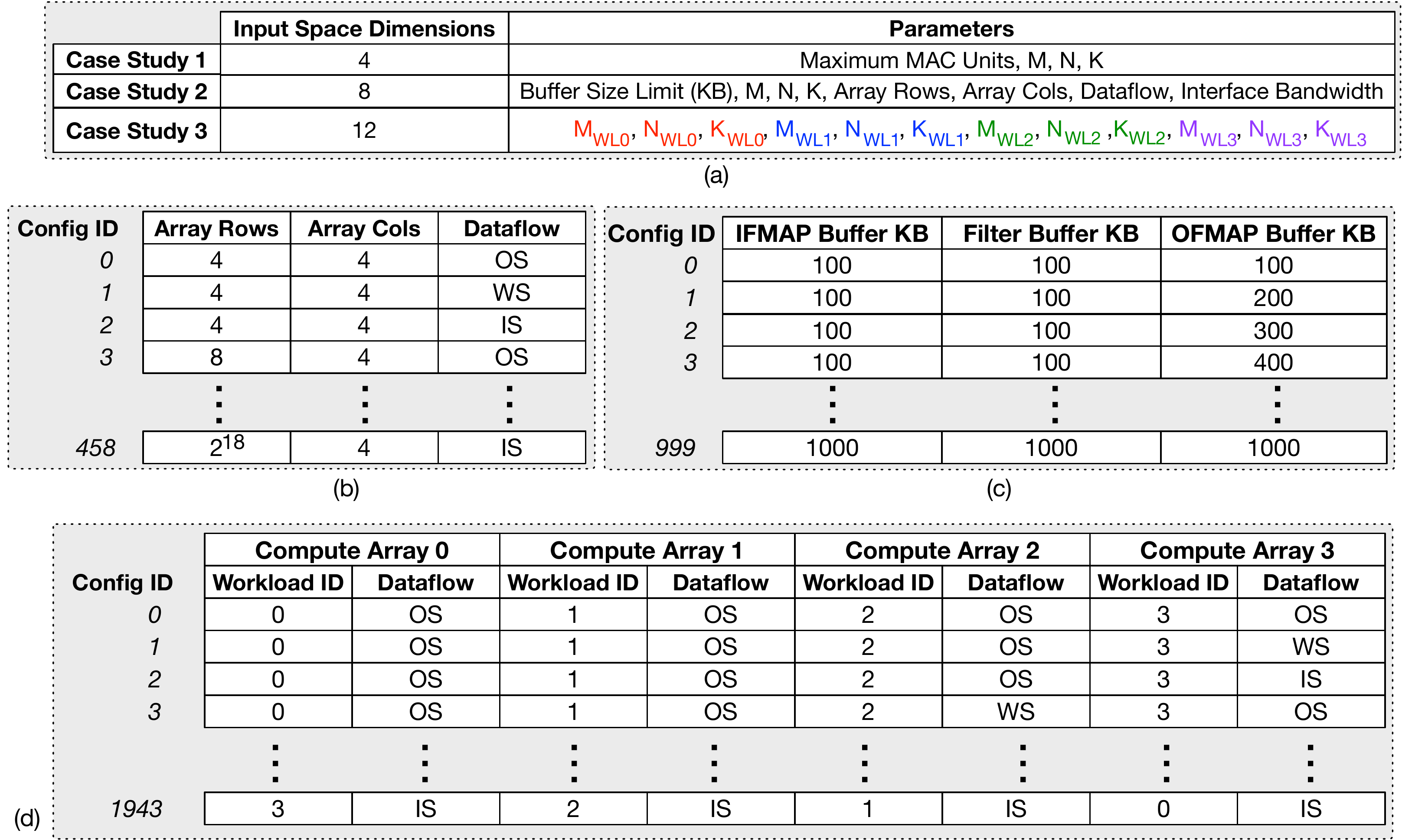}{
    (a) Input space size and input parameters for the three case studies; 
    Output space of
    (b) Systolic Array dimension and dataflow prediction case study
    (c) Memory buffer size prediction case study
    (d) Distributed array schedule prediction case study
}
 
 \textbf{Case Study 1: Array and Dataflow prediction}.
 For this case study, we want to predict the optimal dimensions of a systolic array and the dataflow, given then design constraints and the dimensions of the GEMM workload. 
 \textit{The input space} our for this task, therefore comprises of four integers, three for the operand dimensions (M, N, and K), and one integer capturing the maximum number of MAC units allowed. 
 To keep the input space bounded, the MAC units are provided in exponents of 2. 
 We limit the maximum possible MAC units at $2^{18}$.
The workload dimensions are provided as randomly sampled integers from a uniform distribution. 
 The limits of the workload dimensions are determined by the distribution depicted in \autoref{fig:mnk_distribution_sched_space_growth_comb}(a).
 This distribution is generated from the layer dimensions of popular convolution neural networks \cite{alexnet, mobilenet, googlenet, resnet} . 
 This distribution dictates that the values for M dimension vary between 1 to $10^5$, N in [1, $10^4$], and K in [1, $10^3$].
 \textit{The output space} is a list of labels for this task. 
 As shown in \autoref{fig:input-output-space-tables}(b) each label serves as the index for a set of parameters, which for this case study denote the systolic array height, width, and one of the dataflows. 
 Keeping with the conventional systolic array designs, we only consider the array dimensions that are powers of 2. 
 In our experiments, we allow the minimum array dimensions to be $2^4$ while the maximum dimensions, dictated by the maximum possible MAC units in the input space, come out to be $2^{16}$.
 With these limits in place, the output space for our problem contains 459 different array and mapping configurations.
 To generate the dataset, we use runtime as the cost metric.
 We use the SCALE-Sim \cite{scalesim-arxiv} simulator to generate runtimes for each workload for a given array dimension and workload. 
 We modify the simulator to generate only compute runtime and ignore stalls, speeding up the search. 
 We exhaustively search through all the valid configurations to generate the label.
 We generated about 5 million data points, which took about a couple of weeks of times when running over several servers using a total of 800 logical CPU cores.
 
 
\textbf{Case Study 2: Buffer Size prediction.}
In the second case study, our goal is to predict all optimum sizes of the three operand buffers in the systolic array (see \autoref{fig:SystolicHighLevel}) simultaneously for given workload parameters, information about the compute, and design constraints.

\textit{The input space,} as depicted in \autoref{fig:input-output-space-tables}(a), captures the maximum memory capacity, workload dimensions (M, N, K), the
array dimensions, dataflow, and interface bandwidth of the buffers.
For simplicity, we assume the same interface bandwidth for all three buffers.
The limit for the maximum memory capacity is set to 3000 KB; 
the workload dimensions are sampled from the same distribution as the previous case study with limits observed in \autoref{fig:mnk_distribution_sched_space_growth_comb}(a).
The array dimensions and dataflow are sampled from the output space of the previous case study to keep our experiments consistent.
The interface bandwidth expressed in bytes/cycles is taken from the space of integers in [1, 100], sampled with uniform probability.
\textit{The output space,} is a list of labels, where each label indexes an entry of memory sizes for each of the buffers.
The minimum size of each buffer is restricted to 100 KB, and the allowable sizes increment with a step size of 100 KB as well. The maximum allowable size is 1 MB for an individual buffer.
With these constraints, the output space contains 1000 distinct points defining the search space.
For this case study, the optimal memory size is the one that leads to a minimum or no stall. 
For different configurations with the same stalls encountered, the one with the smallest cumulative capacity is considered the best. 
For this case study as well, we use SCALE-Sim\cite{scalesim-ispass} to generate costs of the various memory sizes. 
We generate about 5 million datapoints, where a million point takes about a week when multiple runs are launched onto a server cluster totaling 800 logical CPUs.

\textbf{Case Study 3: Multi-Array Scheduling}.
In \autoref{subsec:casestudy-sched-space} we describe the problem of scheduling a set of GEMM workloads on an equal number of a distinct heterogeneous collection of compute units.
\textit{The input space}, for this problem, is simply the GEMM workload dimensions (M, N, K) one for each compute array. 
\textit{The output space}, is an id depicting the mapping of workload to the arrays and the corresponding dataflow to be used as shown in \autoref{fig:sched_critical_path_runtime}(a).

\textit{It is worth noting that our case study concerns the schedules when the number and the dimensions of the arrays are fixed. 
This case study does not cover the cases where the number of arrays or the workloads changes.}
However, it is interesting to observe that when the number of arrays changes the possible number of scheduling strategies grows exponentially as well as combinatorially as depicted by the equation $N = 3^{x} \times x!$, where x is the number of compute arrays.
%

\autoref{fig:mnk_distribution_sched_space_growth_comb}(b) depicts the growth of the space with the number of compute units.
\textit{In this case study, we chose to learn the scheduling space of \textbf{four} arrays, which leads to 1944 possible entries in the output space}.
The decision to chose this configuration is purely pragmatic, intended to keep the dataset generation times bounded. 
The different arrays used in the study is shown in \autoref{fig:input-output-space-tables}(c). 

Run time is used as the cost function for dataset generation. 
The schedule which leads to the least runtime of the slowest running workload is deemed to be the winner. 
For the schedules which have the same critical path runtime, the one with the least cumulative runtime is chosen to be optimal.
We create an in-house simulator similar to \cite{scalesim-ispass} to compute the runtime of GEMM workloads on a distributed systolic array setting.
Similar to the previous case studies we generate roughly 5 million data points for learning.

\subsection{Statistical analysis}
\label{subsec:statistical-analysis}

 \vspace{-1mm}



\insertWideFigure{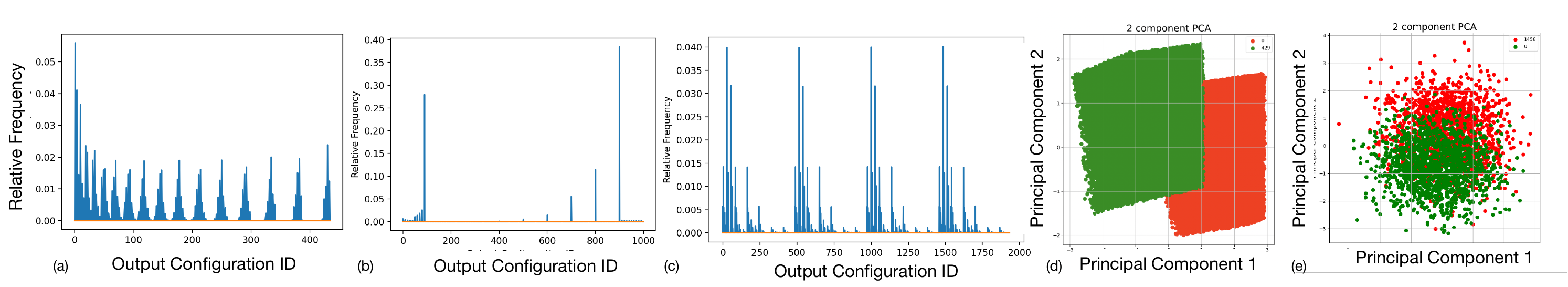}{Distribution of the configuration ids for (a) Case Study 1, (b) Case Study 2, (c) Case Study 3. 
Distribution of datapoints along the most relevant principal component of two configurations for (d) Case Study 1, and (e) Case Study 3.}

In \autoref{sec:design-aware-analysis} we observed that the optimal configurations show distinct patterns when analyzed with manually picked parameters. 
In this section, we perform statistical analysis on the generated datasets to gain additional insights on the ability to learn the distribution. 

First, we analyze the distribution of the categories in the training dataset. 
In \autoref{fig:dataset_distr_pca_comb}(a-c) we plot the relative frequencies of the label categories for three case studies. 
The dataset for our first case study (see \autoref{fig:dataset_distr_pca_comb}(a)), which captures the optimal array dimensions and dataflow for given GEMM workloads, we immediately notice that the output distribution is complicated and non-uniform, but structured. 
We observe that several configurations are represented in the majority while there exist a reasonable amount of configurations that have minuscule representation. 
This is consistent with the observations we make from \autoref{fig:ArrDf-AspectRatio-Scatter-sweep_comb}, where we see that highly skewed array dimensions are optimal only for a small number of use cases, leading to lower frequencies.
It should be noted however that the input space of the dataset, which comprises the dimensions of the GEMM workload and the max number of MAC units is covered uniformly. 
We, therefore, decide to keep the dataset unprocessed and do not force explicit class balancing.
%

In \autoref{fig:dataset_distr_pca_comb}(b) we show the distribution of categories for the second use case, pertaining to the memory sizing problem.
The immediate observation that can be made on this distribution is that for this problem, a handful of configurations work for a wide variety of input configurations, evident by the few classes the dominate the output distribution. The most frequent category accounts for 38\% of the output. The two most frequent capture 66\%. Continuing, the ten most frequent configurations account for roughly 90\% of the output. Beyond this point, other configurations have very few occurrences and are therefore much harder to predict. 
We expect low prediction accuracy on this dataset due to the high bias in the dataset.
%
%

A similar analysis on the scheduling dataset, depicted in \autoref{fig:dataset_distr_pca_comb}(c) reveals a different pattern.
In this case, not only there are a few very prominent configurations, but also a large number of categories with very low frequency. The vast majority of configurations are never optimal and do not occur.  
 We can see that four classes, in particular, dominate the spectrum. Each accounts for roughly 14\% of the output totaling 56\%. There are eight additional classes that each account for roughly 3\% each totaling 24\%. The twelve major classes then account for 80\% of the optimal configurations.
We can take away two key insights from this plot. 
First, a relatively simple model, which can classify the larger spikes can still provide respectable prediction performance. 
However, employing a more sophisticated model can significantly boost the performance of the schedule predictor, if it can learn to classify the labels with smaller frequencies.

Finally, we examine the efficacy of the handcrafted features used in the datasets. 
In \autoref{fig:dataset_distr_pca_comb}(d-e) we plot a subset of the data points along with the most prominent principal components for the first and the third case study respectively for two randomly chosen labels/categories.
The visual separation of the points corresponding to different labels suggest that performing higher dimensional transformation with the chosen input parameters helps to choose a hyper-plane capable of classifying the data points. 
This chart provides a quality assurance that the datasets are classifiable and as a consequence can be learned, although not formal proof.

\subsection{Learning with Existing Classifiers}
\label{subsec:ml-model-search}

 \vspace{-1mm}

\insertFigure{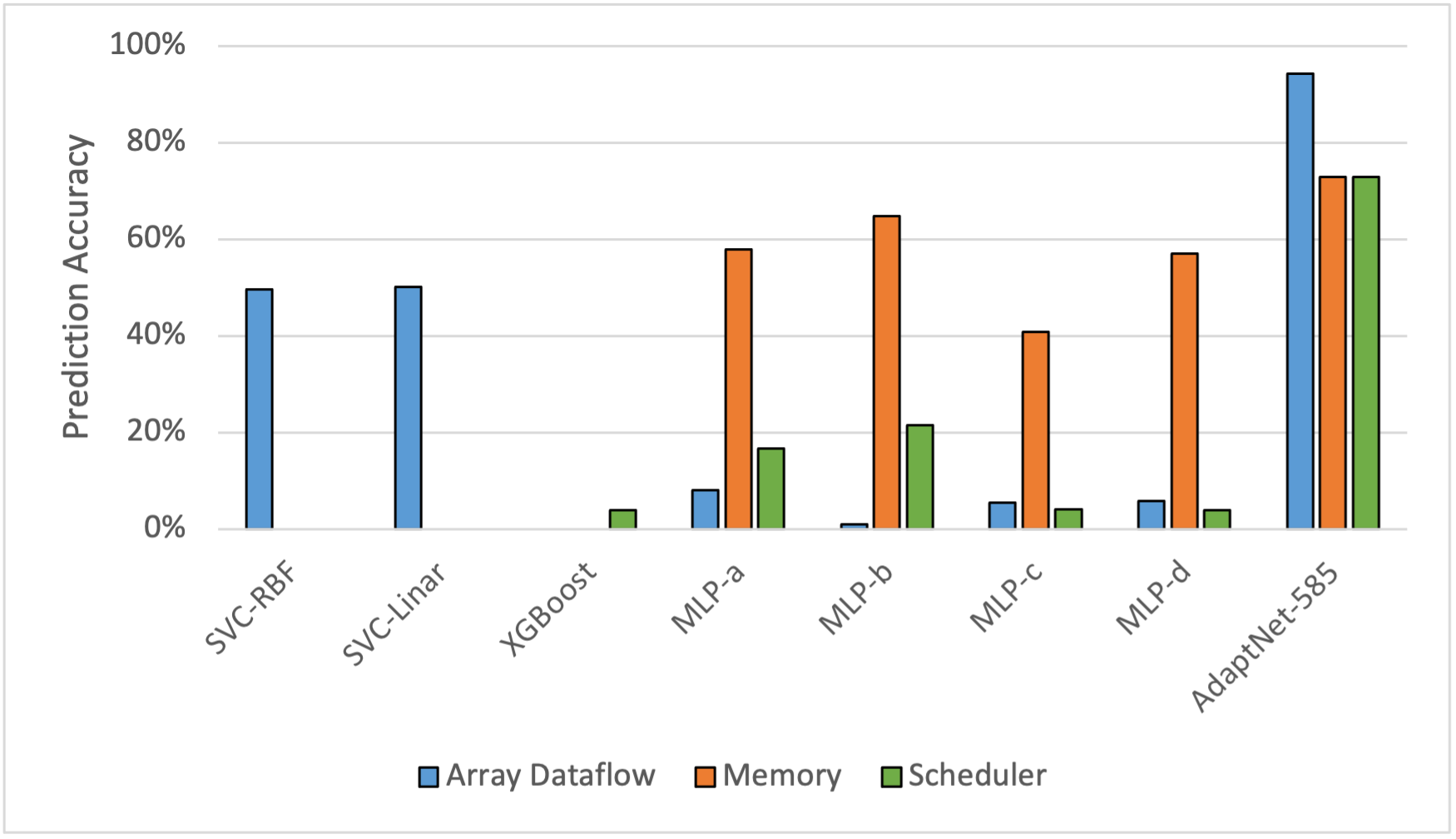}{Performance of Classifier Frameworks. 
}



The problem formulation discussed in \autoref{sec:dataset-generation-analysis}, allows us to use off-the-shelf classifiers to capture the design space and make predictions about the optimal parameters for given workloads and design constraints.
We test out various models with different degrees of complexity on the datasets generated for the three case studies. 
The table in \autoref{fig:other-models} show these models. 
To summarize we used the scikit-learn \cite{sklearn} libraries implementation of support vector classifiers \cite{svm} with linear and radial basis kernel. 
The state-of-the-art tree boosting method called eXtrement gradient boosting (XGBoost \cite{xgboost}) available from the xgboost package \cite{xgboost-package}.
We also implemented four multi-layer perceptron (MLP) networks in TensorFlow's Keras \cite{tensorflow, keras}.

\autoref{fig:other-models} shows the accuracy obtained for the three case studies when $2\times10^{6}$ datapoints are used for fit/training. 
For the MLPs, the networks are trained for 15 epochs with 90:10 training-validation split. 
We used a categorical cross-entropy loss function with accuracy as the optimization objective. 
Among the various case studies, the one for memory size prediction is learned well among all the models, with MLP-B attaining about 63\% validation accuracy. 
Support Vector Classification, however, was able to perform the best among the models considered attaining about 50\% test accuracy for case study 1, 60\% and 40\% for the other two respectively.

\section{\recnet: Design and Analysis}
\label{sec:airchitect}

\vspace{-2mm}

\insertWideFigure{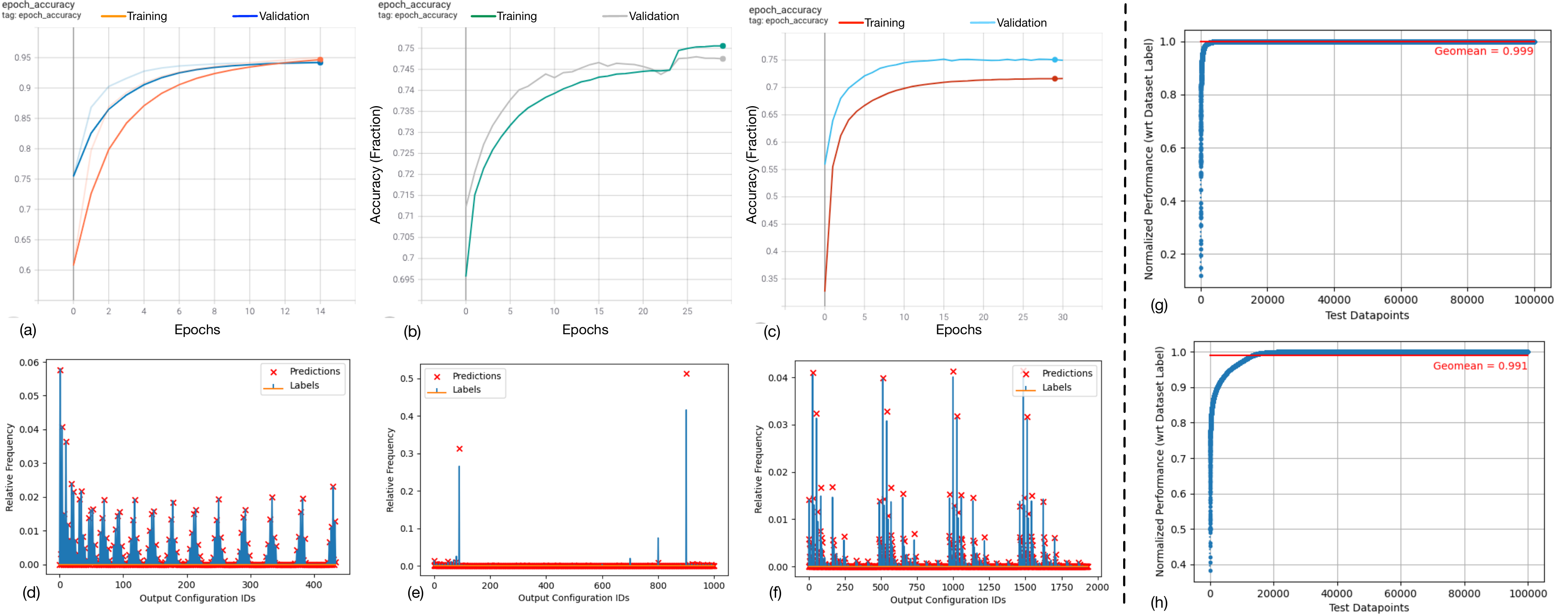}{
    Progression of training and validation accuracy vs epochs when training \recnet on the dataset for (a) Case Study 1, (b) Case Study 2, (c) Case Study 3. 
    The distribution of actual label and predicted configuration IDs by traind \recnet{} on test datasets with $10^5$ points for (d) Case Study 1, (e) Case Study 2, (f) Case Study 3.
    The sorted normalized performance of predicted configurations to the optimal configurations on workloads present in test dataset for (g) Case Study 1, (h) Case Study 3
}

\insertFigure{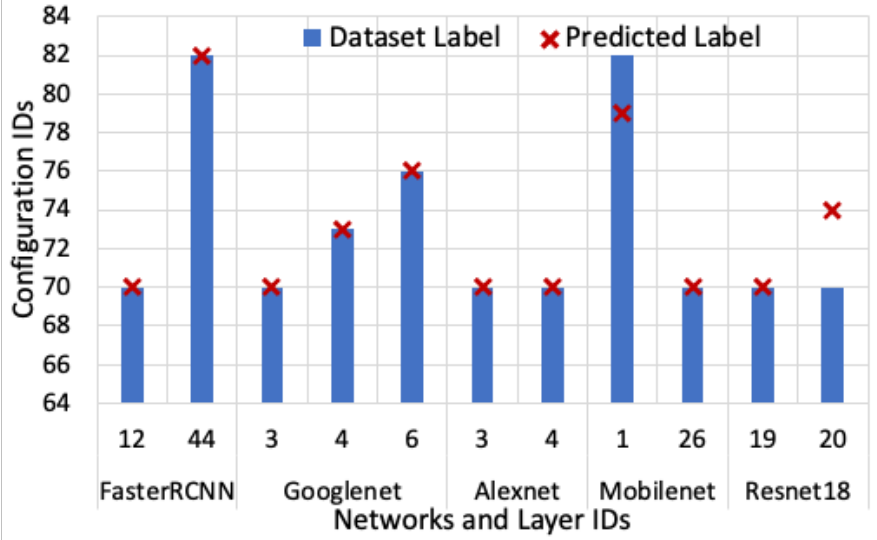}{Predicted and actual labels for case study 1 for a few layers of popular CNNs}


From our analysis in the previous section \autoref{subsec:ml-model-search}, we notice that although off-the-shelf classifiers are capable of learning the design space for our various use cases no single model appears to perform consistently across the different spaces. 
We design a general network structure, which we call \recnet{}, intending to obtain the capability to achieve reasonable learning performance across different distributions.
\autoref{fig:recnet-generic} depicts the general structure of our proposed model. 
The inspiration for this design comes from the structure of the modern recommendation networks like DLRM \cite{dlrm}, which perform recommendation by mapping query inputs onto a trained embedding space, followed by MLP based classification. 
The trained embedding map the input data from the user-defined input space onto a latent space, which immensely improves the performance of the classifiers. 
This is also evident from the performance of MLP-B vs \recnet{} as depicted in \autoref{fig:other-models}.
Among the various case studies, the number of inputs and outputs are the parameters that we changed. 
The number of inputs to the network is equal to the input space, while the network generates a one-hot vector of length equal to the output space of the problem indicating the optimal design or mapping parameters, as discussed in \autoref{subsec:dataset-generation}, \autoref{fig:input-output-space-tables}. 
We use an embedding size of 16 and a 256-node MLP hidden layer for our models across the case studies.

For all three use cases we train the corresponding versions of \recnet{} on the respective datasets with 4.5M points (\autoref{sec:dataset-generation-analysis}) using 90:10 train-validation split. 
We use TensorFlow's Keras library\cite{tensorflow, keras} to implement the network and train using categorical-cross-entropy as the loss function, with accuracy as the optimization metric.
\autoref{fig:tensorboard-outdistr-charts}(a-c) shows the accuracy obtained during training for the three case studies respectively.
We observe that the design space of case study 1 is learned with a high validation accuracy ($>94\%$) in about 15 epochs. 
The training for case study 2 finishes, in about 22 epochs achieving a validation accuracy of 74\% before starting to overfit. 
For case study 3, the network saturates at about 15 epochs at about 76\% validation accuracy.
As we see in \autoref{fig:other-models}, \recnet{} beats the best performing off-the-shelf classifiers at least by about 10\% accuracy.

To gain further insights on the quality of training, we plot the distribution of the actual labels and the predicted configuration IDs for $10^5$ previously unseen test data points for each case study. 
\autoref{fig:tensorboard-outdistr-charts}(d-f) shows predicted vs actual distribution for the three case studies. 
The first observation is that the test datasets' actual distribution closely matches with distributions of the original dataset, shown in \autoref{fig:dataset_distr_pca_comb}(a-c) corroborating that the test set does not compromise on generality.
Second, visually the predicted distribution for case study 1 almost perfectly matches the actual distribution, confirming that the network learned the design space.
Third, we observe that the networks for case study 2 and 3, learn to predict the configurations with significant representation on the dataset, while the configuration IDs with low statistical probability is ignored as noise. 
The presence of large spikes in the case of memory size prediction biases the model, shown by the high frequencies of the top two configuration IDs (\autoref{fig:tensorboard-outdistr-charts}(e)) leading to relatively low accuracy. 
Similarly, for case study 3,\autoref{fig:tensorboard-outdistr-charts}(f) shows that the model was successful in learning the distribution of the large spikes. 
However, in doing so it ignored the configurations with lower frequencies, which in turn cumulatively lowered the accuracy figure. 
\textit{However, it is worth noting that the model ignoring the lower frequencies as statistical noise demonstrates that the model is robust and does not overfit for any of the use cases.}
Further improving prediction accuracy for different design spaces requires data engineering on a case-by-case basis and therefore, is out of the scope of this paper.

To understand the cost of misprediction within the use cases, we compute the performance (reciprocal of runtime) of the workloads in the test dataset, for the predicted and label configurations. 
In \autoref{fig:tensorboard-outdistr-charts}(g) we show the performance of the predicted configurations normalized to the labels for the $10^5$ datapoints. 
Due to the high prediction accuracy, we observe that only a few data points have catastrophic performance losses (<20\% of the optimal) leading to a 99.99\% average performance (Geometric Mean) of the best possible.
In \autoref{fig:ArrDf-RealWL} we depict the performance of the network on some layers from networks like FasterRCNN\cite{fasterrcnn}, GoogleNet\cite{googlenet}, Alexnet\cite{alexnet}, MobileNet\cite{mobilenet} and ResNet-18\cite{resnet}. 
The layers of any of these networks were not part of the training or validation dataset, but the model is able to predict the optimal array shapes and dataflow, when queried with a constraint of $2^{10}$ MAC unit limit.
Interestingly for case study 3, which had relatively low accuracy, for most cases, the performance does not lead to catastrophic losses.
\autoref{fig:tensorboard-outdistr-charts}(h) shows that among the mispredictions, most of the points suffer about 10\% to 15\% loss compared to the best achievable, leading to an average of 99.1\% of the best possible runtime on average (GeoMean).

\section{Related Works}
\label{sec:related-works}

\vspace{-2mm}

\textbf{ML for Architecture search:}
Apollo \cite{apollo} is a recent work from Google, targeting sample efficient searching through the accelerator design space using reinforcement learning.
Gamma\cite{gamma} and ConfuciuX\cite{confuciux} are similar ML based architecture mapping and design space configuration search methods which use genetic algorithm and reinforcement learning (RL) respectively.
AutoTVM\cite{autotvm} use ML model for cost prediction to improve fast mapping search during compile time.

\textbf{ML for EDA:}
Recently there has been a significant push toward automating place-and-route using machine learning.
Mirhoseni et al\cite{mirhoseini2017device} use RL for task placement on a heterogeneous system.
Wang et al\cite{wang2020gcn} use GCN and RL for automatic transistor sizing.
NVCell\cite{nvcell} is a RL based proposal from Nvidia to automate standard cell placement.
Nautilus\cite{nautilus} uses genetic algorithm to improve FPGA place and route.
Kwon et al\cite{kwon2019learning}, use online tensor-based recommender systems to aid place and route in chip design.
\textit{To the best of our knowledge this work is the first work, which truly learns the architecture design and mapping space and "generalizes" to predict the optimal configuration in constant time without the requirement of searching the design space.}
\section{Conclusion}
\label{sec:conclusion}
\vspace{-2mm}

This paper presents \recnet{}, a recommendation neural network for learning the architecture design and mapping space of systolic array based accelerators.
To the best of our knowledge is the first work to show that architecture design space is learneable. 
Our approach significantly reduces the cost of design space exploration, by replacing expensive search with constant time recommendation of optima.
The paper shows a systematic study of the design and mapping space in the context of design aware and statistical analysis, demonstrates the formulation of conventional search problem in to a machine learning problem, and finally proposes a model which outperform the existing off-the-shelf ML models to capture the mapping and design space by significant margin.

\bibliographystyle{IEEEtranS}
\bibliography{references}

\end{document}